\documentclass[lettersize,journal]{IEEEtran}
\usepackage{amsmath,amsfonts}
\usepackage{array}
\usepackage[caption=false,font=normalsize,labelfont=sf,textfont=sf]{subfig}
\usepackage{textcomp}
\usepackage{stfloats}
\usepackage{url}
\usepackage{verbatim}
\usepackage{graphicx}

\usepackage[utf8]{inputenc} 
\usepackage[T1]{fontenc}    
\usepackage{hyperref,soul}       
\usepackage{url}            
\usepackage{booktabs}       
\usepackage{amsfonts}       
\usepackage{nicefrac}       
\usepackage{microtype}      
\usepackage{xcolor}         
\usepackage{algorithm}
\usepackage{algpseudocode}
\usepackage{cite}

\usepackage{amsmath, amssymb, graphicx}
\usepackage{caption}

\def\BibTeX{{\rm B\kern-.05em{\sc i\kern-.025em b}\kern-.08em
    T\kern-.1667em\lower.7ex\hbox{E}\kern-.125emX}}
\usepackage{balance}
\begin{document}
\title{Context-Matched  Collage Generation for Underwater Invertebrate Detection}
\author{R. Austin McEver, Bowen Zhang, B.S. Manjunath
\thanks{Austin McEver, Bowen Zhang, and B.S. Manjunath are with University of California, Santa Barbara, 93106, California, United States of America. Email: mcever@ucsb.edu; bowen68@ucsb.edu; manj@ucsb.edu;}}%


\maketitle

\begin{abstract}
The quality and size of training sets often limit the performance of many state of the art object detectors. However, in many scenarios, it can be difficult to collect images for training, not to mention the costs associated with collecting annotations suitable for training these object detectors. For these reasons, on challenging video datasets such as the Dataset for Underwater Substrate and Invertebrate Analysis (DUSIA), budgets may only allow for collecting and providing partial annotations \cite{mcever2022context}. To aid in the challenges associated with training with limited and partial annotations, we introduce Context Matched Collages, which leverage explicit context labels to combine unused background examples with existing annotated data to synthesize additional training samples that ultimately improve object detection performance. By combining a set of our generated collage images with the original training set, we see improved performance using three different object detectors on DUSIA, ultimately achieving state of the art object detection performance on the dataset.
\end{abstract}


\section{Introduction}
\IEEEPARstart{T}oday's computer vision methods largely depend on enormous datasets with many annotated examples for each class. These sorts of datasets can be extremely expensive to collect, especially when the data is more scientific in nature. While any layperson can label a cat, dog, or human, the cost of labelling and differentiating between more specific, scientific classes grow exponentially. The cost of collecting, annotating, and analyzing scientific data is high, but that also means that any automation or streamlining of those processes can greatly benefit domain scientists who currently rely almost entirely on expensive human experts.

\begin{figure}
    \centering
    \includegraphics[width=0.85\linewidth]{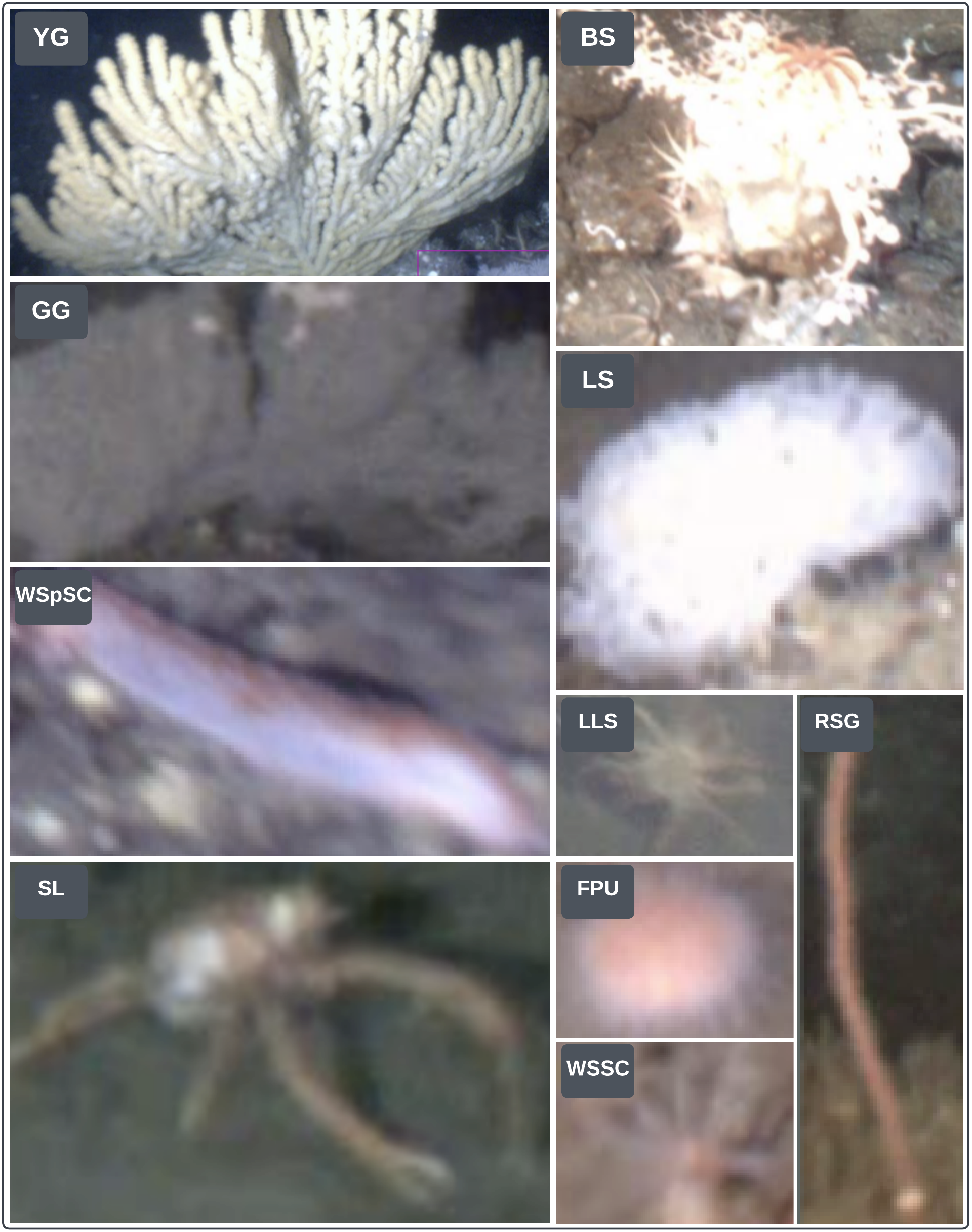}
    \caption{Examples of the ten species that are labelled with bounding boxes in DUSIA. YG stands for yellow gorgonian; BS, basket star; GG, gray gorgonian; LS, laced sponge; WSpSC, white spine sea cucumber; LLS, long-legged sunflower star; SL, squat lobster; FPU, fragile pink urchin; WSSC, white sea slipper cucumber; RSG, red swifita gorgonian.}
    \label{fig:dus_species}
\end{figure}

The Dataset for Underwater Substrate and Invertebrate Analysis (DUSIA) \cite{mcever2022context} provides an example of a challenging, scientific dataset. DUSIA contains 10 hours of video collected in 1080p using a remotely operated vehicle (ROV) that drives over and records the ocean floor at depths between 100 m and 400 m, and the data within DUSIA is part of a much greater, growing collection of hundreds of hours of unlabelled or weakly labelled videos. Marine scientists collect these videos as part of surveys that improve their understanding of habitats and organisms of the ocean floor. DUSIA's videos come directly from marine scientists working to study, understand, and survey the ocean floor.

Despite the rich content of the videos, DUSIA's annotations are limited due to the expense of hiring trained marine science experts to annotate video with the level of granularity of typical computer vision datasets. The dataset provides numerous, weak labels, which indicate timestamps at which 57 invertebrate species of interest are Counted At the Bottom Of the video Frame (CABOF), as well as a training set with frames partially annotated with bounding boxes for the the ten species shown in Figure \ref{fig:dus_species}. 

\begin{figure*}
    \centering
    \includegraphics[width=1.0\linewidth]{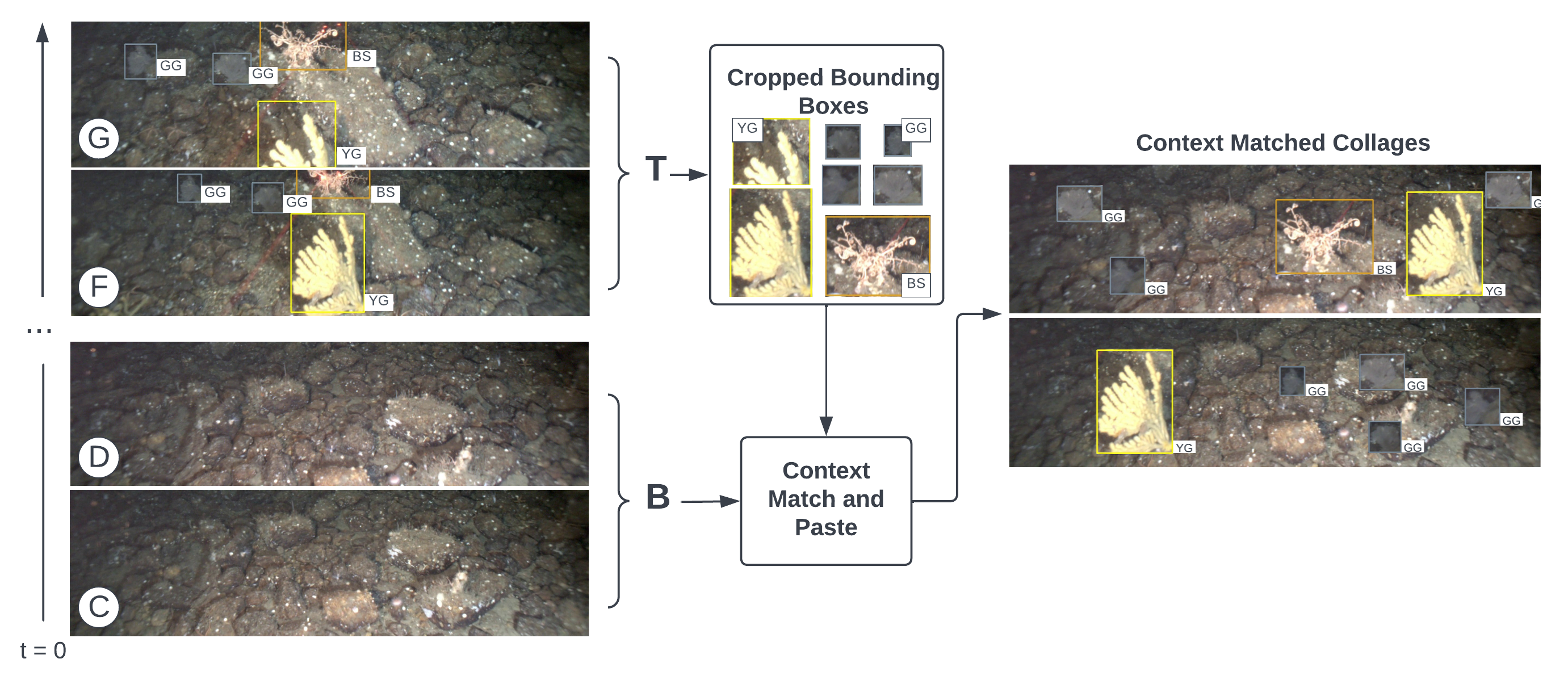}
    \caption{Diagram illustrating the method for generating Context Matched Collages. Mine bounding boxes from training set $\mathbf{T}$, background frames from $\mathbf{B}$, match the context, and paste boxes on to a context matched frame from $\mathbf{B}$. See Figure \ref{fig:dus_species} caption for species name abbreviations.}
    \label{fig:collagemethod}
\end{figure*}

CABOF labels are described in detail in the original work \cite{mcever2022context} and illustrated via a frame by frame representation of DUSIA's videos in Figure \ref{fig:collagemethod}. In summary, as the ROV traverses the ocean floor, species come into view at the top of the frame and make their way to the bottom of the frame as the ROV and video moves forward. Cropped example frames are shown in Figure \ref{fig:collagemethod} with frames going forward in time from bottom to top. When a species individual first touches the bottom of the frame (like the yellow gorgonian in Frame F of Figure \ref{fig:collagemethod}), annotators create a CABOF label with that species name and the timestamp, which corresponds to collected GPS coordinates. This labelling gives marine science researchers a metric for counting the number of species individuals occurring along a narrow transect path. In Section \ref{sec:method}, we present a new use for these CABOF labels and leverage them to try to find frames in DUSIA's videos where there are \emph{no} species.

DUSIA presents partial bounding box labels for training because collecting full labels is preventatively expensive. These labels are partial in that every instance of a species of interest in the training set's frames may not be annotated. That is, there may be some unlabelled individuals of species of interest in the training frames. 

DUSIA's partial annotations provide an interesting challenge for today's computer vision methods and require new methods to solve the object detection problems presented by the dataset. While unconventional, using computer vision on challenging, scientific datasets opens up new possibilities for computer vision applications. One new possibility may include generating synthetic data to supplement and enhance small, noisy training sets.

Our contributions are as follows: 
\begin{itemize}
    \item Our method leverages explicit context labels available in DUSIA to generate new training samples that combine existing training samples with empty background frames available in sparsely populated video areas, illustrating that cutting bounding boxes from the training set (as opposed to cutting more precise, segmented class instances) can serve as an effective basis for data augmentation.
    \item We introduce a computationally inexpensive method for leveraging DUSIA's CABOF labels for generating effective training samples achieving state of the art detection results on DUSIA's validation and test sets using multiple different popular object detection models. 
\end{itemize}

\section{Related Work}

Computer vision researchers have long been aware of the power of data augmentation methods for improving training object detectors and image classifiers, and recently, much work has gone into generating plausible training samples via cut/paste methods. Cut/paste methods take objects of interest, cut them from their original image, and paste them into another training image or other type of canvas (e.g. a blank background).
DeVrires et al. \cite{devries2017improved} introduce Cutout as a method of cutting portions from images during training to help improve performance in the image classification task, and CutMix \cite{yun2019cutmix} builds upon Cutout by combining two training samples at a time by cropping a random part of one image and pasting it on to another image.

Cut, Paste, and Learn \cite{dwibedi2017cut} leverages separate collections of common images of objects and typical indoor scene examples to cut object instances from the images available for training and paste them to random background scene images. Cutting object instances from images relies on an image segmentation model to separate objects from their backgrounds, and then those instances are randomly pasted on random indoor images.


Ghiasi et al. \cite{ghiasi2021simple} perform an augmentation similar to Cut/Paste, and Learn where they cut object instances from one image to paste on to a different, randomly selected image. They consider indoor vs outdoor images, which they label based on COCO's panoptic labels. They then use these augmented images to train an image segmentation model. In the medical domain,  TumorCP \cite{yang2021tumorcp} leverages image segmentation labels to create additional training samples for a segmentation network, leading to better segmentation performance.

ObjectMix leverages instance segmentation labels to augment data for action recognition in videos by extracting object segments from two videos and combining them to create new video samples \cite{kimata2022objectmix}. Similarly, Continuous Copy-Paste works to leverage instance segmentation labels to generate training samples for training models to solve the Multi-object Tracking problem \cite{Xu2021CCP}.

Dvornik et al. show the importance of context in cut paste methods \cite{dvornik2018modeling}. Their method involves training a network using both bounding box and instance segmentation labels to generate a notion of context where the bounding box includes pixels that are not included by the segmentation label. They train a network to then predict possible paste locations for cut out object instances so that objects are pasted on to images that their model predicts to be sensible.

Our method also employs a cut-paste based method and illustrates the importance of context in our scenario but in a few key different ways. For one, our method does not rely on any expensive segmentation labels, which label every pixel in an image, or segmentation models, which may segment objects unreliably. Section \ref{sec:expers} shows that directly cutting a whole bounding box labelled as an object of interest from one frame and pasting it to another frame with matching context enables an improvement in object detection performance. This is important because segmentation labels are difficult and expensive to collect in many scenarios with challenging, scientific data like DUSIA \cite{mcever2022context}, on which we present our results.




\section{Generating Context Matched Collages}
\label{sec:method}
Our method for generating synthetic frames from existing ones is a simple but powerful extension of typical cut-paste methods. We introduce novel changes to this method that allow us to generate better training samples for our specific DUSIA dataset. DUSIA contains 10 hours of video captured in 1080p at 30 fps, but across all of that video footage only 8,682 partially annotated frames contain bounding box labels suitable for supervising most of today's object detectors. These frames are considered partially annotated because they may contain individuals of species of interest that are not labelled with bounding boxes. 

Still, DUSIA provides CABOF labels for the entirety of its videos. These CABOF labels indicate the first time at which a group or individual of a species of interest intersects with the bottom of a video frame (like the yellow gorgonian shown in Frame F of Figure \ref{fig:collagemethod}) such that there is a single CABOF label for every single individual of a species of interest that touches the bottom of a video frame in DUSIA's videos. Our method aims to leverage these CABOF labels in a unique way.

\begin{figure*}
    \centering
    \begin{tabular}{@{}c@{}}
        \includegraphics[width=1.0\linewidth]{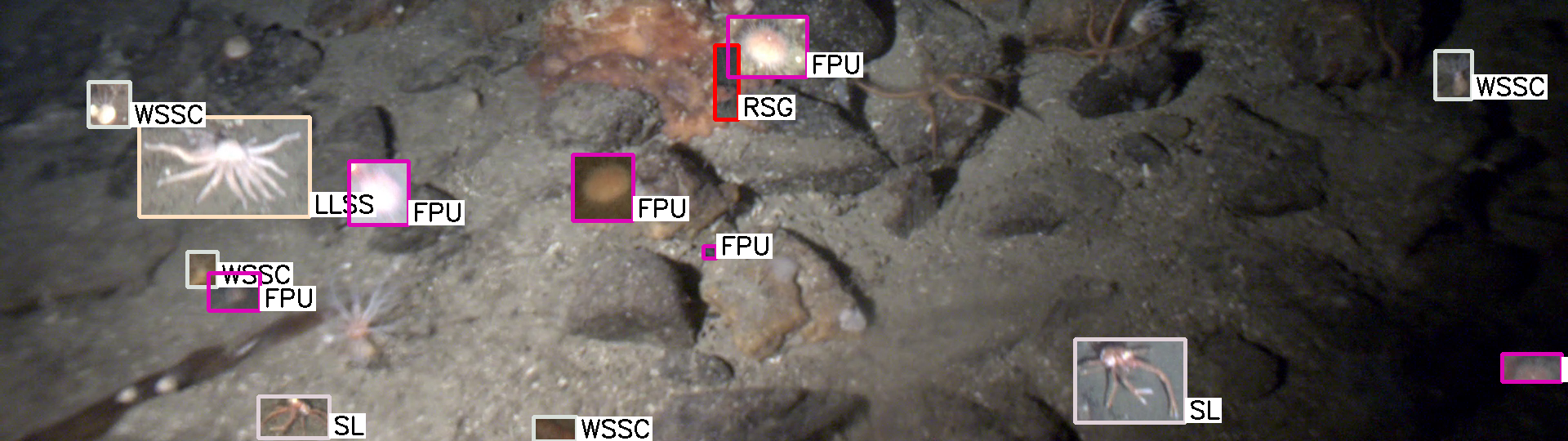}
        \\
        \includegraphics[width=1.0\linewidth]{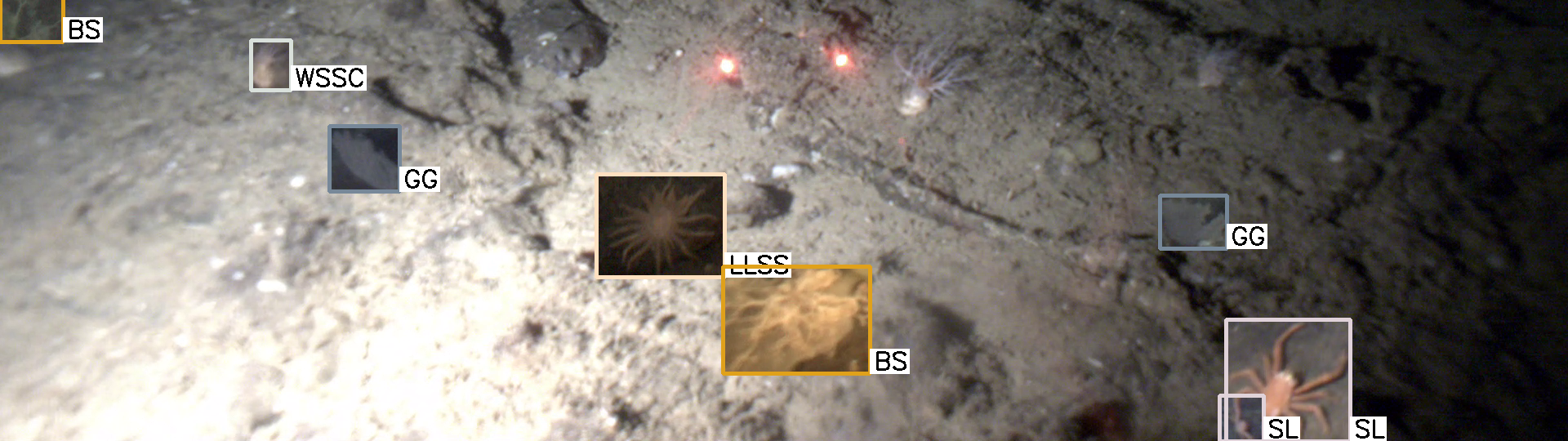}
    \end{tabular}
    \caption{Examples of generated collage images. Top: bounding boxes from images labelled Cobble/Mud pasted on to an empty, background image with the matched Cobble/Mud label. Bottom: images and background labelled Mud. See caption of figure \ref{fig:dus_species} for species abbreviations.}
    \label{fig:collage}
\end{figure*}

Unfortunately, a training set of 8,682 frames limits the performance of the object detectors, but collecting additional bounding box annotations is expensive, especially given the challenging nature of DUSIA and its object classes. DUSIA's partial annotation scheme alleviates the annotation burden on expensive expert annotators, but the scheme makes it very difficult to train an effective object detection network. McEver et al. \cite{mcever2022context} demonstrate that Negative Region Dropping (NRD) can help train Faster RCNN based models on the partially annotated dataset, but the detector that they present is far from perfect. We propose combining DUSIA's original training set with Context Matched Collages as a complementary method that enables even better detection performance.

The first step in generating these collages is to find a set of frames from the videos that can serve as background for the synthetic collage training samples. Ideally, these background frames contain a minimal number of species individuals so that the resulting collages can have few unlabelled species individuals in them. Pasting on to these sort of background frames allows the object detector to see more of the video, helping it generalize on the test set. Further, by pasting known objects on to empty frames, we can generate frames that are better supervised than some of the partially annotated training set, as they contain fewer unlabelled species of interest.

To this end, we generate a set of frames, $\mathbf{B}$, that are unlikely to contain species of interest. In order to do so, we leverage the Count at Bottom of Frame (CABOF) labels, which indicate timestamps containing  species individuals, provided for DUSIA's videos. We can therefore use frames that are far from all CABOF labels to find all the time spans that are unlikely to contain a species of interest. 

We first initialize $\mathbf{B}$ to contain all frames in the training set. Then, we iterate through all CABOF labels removing frames within a certain range (e.g. a few seconds) of any CABOF label time stamp. For example, if a CABOF label indicates that there are three fragile pink urchins at time 00:10:30, we can remove all frames ranging from 00:10:20 to 00:10:40 from $\mathbf{B}$. Because not all species individuals that appear in the video touch the bottom of the video frame, they do not all get a CABOF label, but many individuals do. Additionally, since many species typically occur together, this step helps filter out very busy parts of the video with lots of species of interest in them, regardless of whether they touch the bottom of the frame because it is likely that their neighboring species individuals touch the bottom of the frame if they do not. The remaining frames in $\mathbf{B}$ may contain a few unlabelled species of interest, but there should be far fewer unlabelled species than in the original training frames, which are known to be partially supervised and typically come from busy parts of the videos that contain many intermingling groups of species of interest. 

Given that nearly all frames in DUSIA have substrate labels that indicate the substrate present on the ocean floor, we  also create a map $\mathbf{S}$ that maps each substrate combination, $s$ to the frames from $\mathbf{B}$ that contain that substrate combination. Having this mapping allows us to match the context (i.e. substrate label) of potential background frames and the context of any bounding boxes we wish to paste into a new collage frame.

In order to generate the final collages, we cut bounding boxes from the original training set and paste them on to images from $\mathbf{B}$ that have matching substrate labels. To do so, we map all substrate combinations to a list of bounding boxes that exist on frames with that substrate combination label. For each substrate combination, we randomly sample boxes, and paste them to random locations in a randomly selected image from $\mathbf{B}$ that has matching context labels. In our case we randomly select between 1 and 15 boxes to paste on each image. We also ensure that boxes do not fully occlude one another, though we do allow partial overlap because species individuals often cluster closely together in the original videos.

While the generated collage images may be obviously manipulated to a human eye, they help train a stronger object detection model by providing better supervision, unique co-occurrences of species individuals, and more samples. We explore these improvements in Section \ref{sec:expers}

\begin{algorithm}[]
\caption{Pseudo code for generating Context Matched Collages}\label{alg:cap}
\begin{algorithmic}
\State $b \gets$ 10 seconds \Comment{buffer}
\State $B \gets$ all training frames
\State $C \gets$ all CABOF labels
\For{$c \in C$} \Comment{Create set of potential background frames}
\State $T \gets c.timestamp$
\State $B.remove([T - b, T + b])$
\EndFor

\State $L \gets$ map of each substrate label to empty list
\For {$f \in B$} \Comment{map substrate combos to bg frames}
\State $L[f.substrate].append(f.timestamp)$ 
\EndFor

\State $K \gets$ map of each substrate label to bounding box labels
\State $M \gets \emptyset$ \Comment{list of generated frames}
\While {not done}
\\ \Comment{k is substrate label, O is list of boxes on substrate k}
\For {$k,O \in K.items()$}
\State $l \gets L[k]$ \Comment{list of bg w/ label k}
\State $m \gets random.choice(l)$ \Comment{m can serve as bg}
\State $r \gets random.randint([1,MAX\_BOXES])$
\For{$i \in [0,r)$}
\State $o \gets random.choice(O)$
\State $O.remove(o)$
\State $m.paste(o)$ \Comment{paste box o randomly on bg}
\EndFor
\State $M.append(m)$ \Comment{add Context Matched Collage}
\If{$len(M)$ > MIN} 
\State $done \gets True$
\State break 
\EndIf
\EndFor
\EndWhile

\end{algorithmic}
\end{algorithm}

Figure \ref{fig:collage} shows example Context Matched Collages generated by our method.
\section{Experiments}
\label{sec:expers}
We train multiple model architectures by combining DUSIA's 8,682 training frames with collages generated via our method. We refer to those 8,682 training frames as $\mathbf{T}$. We paste a maximum of 15 boxes onto each background image. We use a buffer of 10 seconds around each CABOF label to ensure that our background images are sufficiently empty.

In order to illustrate the importance of context matching, we generate two collage sets. $\mathbf{M}$ contains approximately 2,000 frames generated as described in Section \ref{sec:method} with contexts matched properly. In practice, there are a small number of bounding box labels with substrate combinations not present in $\mathbf{B}$. For those frames, we simply map to the nearest substrate combination. For example, if a box's substrate label is both mud and cobble, we paste it to a background frame that is either mud or cobble if there are no frames in $\mathbf{B}$ with the exact same substrate label of mud and cobble.

The second collage set, $\mathbf{R}$, contains approximately 2,000 frames generated in exactly the same way as $\mathbf{M}$ except the context of the background frame and the pasted object frame are \emph{not necessarily} matched. That is, the background frame is randomly selected from all of $\mathbf{B}$ rather than randomly selected from a subset of $\mathbf{B}$ with the matched substrate combination.

When training Faster R-CNN with negative region dropping, we saw success in first training on $\mathbf{T}+\mathbf{M}$ then lowering the learning rate to finetune on $\mathbf{T}$. We also present results on that setting indicated by $\mathbf{T}+\mathbf{M}, \mathbf{T}$.

First, we tested the Context-Driven Detector (CDD) as proposed by McEver et al. \cite{mcever2022context}; however, we found that the detector performed better without the context description branch when trained with our collage augmented training sets. We theorize this may be due to the imperfect matching of context in some frames. Still, we found negative region dropping to help overall performance, and we set the negative region dropping percentage, $\rho$ to 0.75 as in the original paper, and, also following \cite{mcever2022context}, we used a learning rate of 0.01 for Faster R-CNN with Negative Region Dropping. 

We also trained YOLOv5 \cite{jocher2020ultralytics} on each of our training sets to demonstrate the impact of including our Context Matched Collages. We trained the YOLOv5 large model from the author's provided weights, which were pre-trained on the COCO dataset \cite{lin2014coco}. After testing a variety of different hyperparameter settings, we found the best performance with most of the default settings; however, we changed the initial learning rate, $lr_0$, from 0.001 to 0.0001; the final OneCycleLR \cite{smith2019super} learning rate, $lr_f$, from 0.1 to 0.01; the anchor-multiple threshold \cite{jocher2020yolov5}, $anchor_t$, from 4.0 to 2.0; the image rotation, $degrees$, from 0.0 to 30.0; and the image perspective \cite{jocher2020yolov5}, $perspective$, from 0.0 to 0.001. After training using the above settings, we finetuned on $\mathbf{T}$ lowering $lr_0$ to 1e-6. 

Finally, we tested the DEtection TRansformer (DETR) \cite{carion2020end} to give an additional example of a state of the art object detection framework. Starting from the author's provided pretrained weights, we trained DETR with minor changes to the default settings. We used ResNet-50 as the backbone and changed learning rate to 1e-5 and the learning rate drop to 40 epochs.

Table \ref{tab:mAPs} shows the experimental results of all three detectors trained on the different training sets without any collages, including the Context Matched Collages, and the collages with random backgrounds. We evaluate our detection results using mean Average Precision (mAP) with an intersection over union (IOU) threshold of 0.5 following \cite{mcever2022context}. We choose this metric, widely known as AP$_{50}$, because the detection tasks in DUSIA is not sensitive to exact localization as the detection tasks ultimately aim to aid in counting of invertebrate species. For more in depth analysis, we also present the popular full COCO suite \cite{zhao2019object} of evaluation metrics for our best detection model, Faster R-CNN with Negative Region Dropping trained on $\mathbf{T}+\mathbf{M}, \mathbf{T}$ in Table \ref{tab:coco}.

\begin{figure*}
    \centering
    \includegraphics[width=0.84\linewidth]{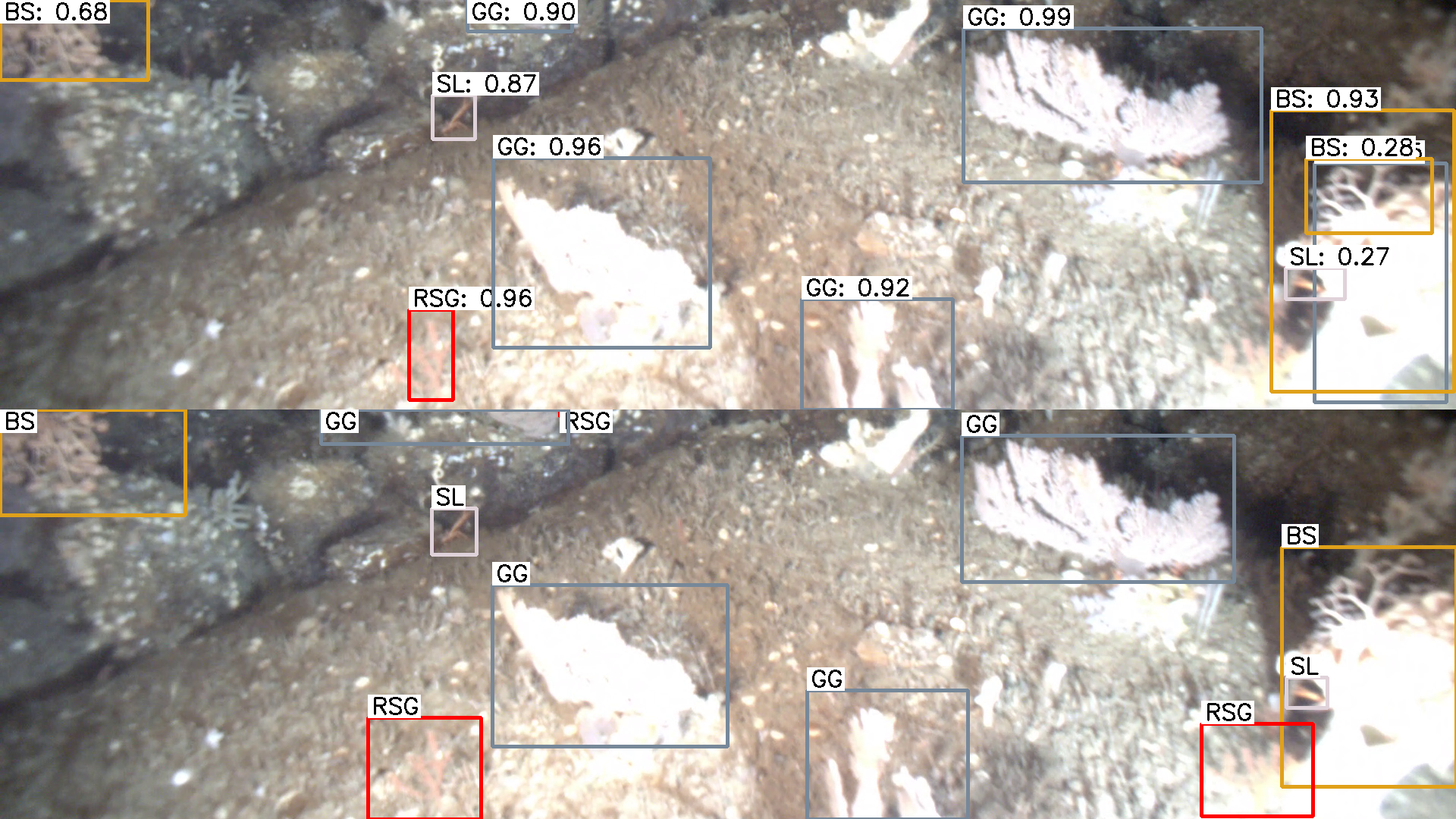}
    \caption{Detections from best model (top) and ground truth (bottom) for an example frame. See figure \ref{fig:dus_species} caption for species name abbreviations.}
    \label{fig:detections}
\end{figure*}

For our best model we also present some qualitative detection results in Figure \ref{fig:detections}. The results for this busy frame show that the detector performs quite well at finding objects at the edge of the frame, and it does a good job discriminating among species. It even leaves out a few sponge species that are not species of interest. The detector still struggles with the red swifitia gorgonian (RSG) class, which is one of the most challenging in DUSIA, showing some area for improvement.

\begin{table}[t]
\centering
\begin{tabular}{@{}llrr@{}}
\toprule
\textbf{Detector}                                 & \textbf{Train Set} & \multicolumn{1}{l}{\textbf{val mAP}} & \multicolumn{1}{l}{\textbf{test mAP}} \\ \midrule
Context Driven Detector \cite{mcever2022context} & T                  & 0.524                                & 0.447                                 \\ \midrule
DETR \cite{carion2020end}                                              & T                  & 0.534                                & 0.416                                 \\
DETR                                              & T+R                & 0.541                                & 0.426                                 \\
DETR                                              & T+M                & 0.541                                & 0.446                                 \\ \midrule
YOLOv5 \cite{jocher2020ultralytics}                                           & T                  & 0.558                                & 0.452                                 \\
YOLOv5                                            & T+R                & 0.518                                & 0.437                                 \\
YOLOv5                                            & T+M                & \textbf{0.558}                       & 0.470                                 \\ \midrule
Faster RCNN \cite{ren2015faster} w/ NRD \cite{mcever2022context}                              & T                  & 0.509                                & 0.439                                 \\
Faster RCNN w/ NRD                                & T+R                & 0.511                                & 0.419                                 \\
Faster RCNN w/ NRD                                & T+M                & 0.542                                & 0.453                                 \\
Faster RCNN w/ NRD                                & T+M, T             & 0.546                                & \textbf{0.482}                        \\ \bottomrule
\end{tabular}
\caption{Different object detectors, and their detection performance given different training sets. CDD results from \cite{mcever2022context}}
\label{tab:mAPs}
\end{table}

\begin{table}[]
\centering
\begin{tabular}{@{}lrr@{}}
\toprule
metric  & \multicolumn{1}{l}{val} & \multicolumn{1}{l}{test} \\ \midrule
AP$_{50:95}$ & 0.264                   & 0.221                    \\
AP$_{50}$    & 0.553                   & 0.482                    \\
AP$_{75}$    & 0.224                   & 0.174                    \\
AP$_{S}$    & 0.017                   & 0.016                    \\
AP$_{M}$     & 0.168                   & 0.165                    \\
AP$_{L}$     & 0.335                   & 0.286                    \\ \bottomrule
\end{tabular}
\caption{Full COCO suite of metrics showing the performance of the best Faster R-CNN with NRD model on both the val and test sets}
\label{tab:coco}
\end{table}

For all three detectors, we see better generalizability in the model, as evidenced by test mAP, when the model is trained with Context Matched Collages. We also see that training with Context Matched Collages ($\mathbf{T}+\mathbf{M}$) consistently achieves better performance than training with the the collages without context matching ($\mathbf{T}+\mathbf{R}$). YOLOv5 and Faster RCNN even see decreased performance when trained with collages in $\mathbf{R}$ illustrating the importance of context when detecting invertebrate species. Faster R-CNN achieves state of the art performance on the test set when trained with Context Matched Collages after finetuning on the original training set. Clearly, augmenting DUSIA's training set with Context Matched Collages leads to better overall performance.

\section{Discussion}
In this paper, we introduce Context Matched Collages. We mine frames containing few species of interest, cut bounding boxes from our training set, and paste those bounding boxes on to the mined images. This process leverages many unused video frames and produces unique training samples that aid in training object detectors to increase performance. We illustrate that cutting bounding boxes, as opposed to finely segmented object instances, and pasting them to create new training samples provides an effective augmentation for object detectors. Further, we introduce matching the explicit context labels of bounding boxes and the background to create collages. By augmenting the original training set of DUSIA with these Context Matched Collages, we are able to achieve state of the art object detection performance. 

Even with these improvements, detection on DUSIA remains a challenging task, and the detection performance still needs improvement to alleviate the manual detection and counting of invertebrate species. The low performance on small objects, shown as AP$_S$ in Table \ref{tab:coco} shows reveals an area for improvement, and the qualitative results in Figure \ref{fig:detections} show that measures may be taken to improve the performance on the red swiftia gorgonian class and perhaps other specific classes. 

\section{Acknowledgements}
This research was supported in part by National Science Foundation (NSF) award: SSI \# 1664172. We would like to thank Dirk Rosen and Andy Lauermann from Marine Applied Research \& Exploration group for their video collection, guidance, and help through this project. We would also like to thank Dr. Robert Miller for their contributions to the project.

\bibliographystyle{IEEEtran}
\bibliography{main}

\end{document}


\onecolumn
\title{Supplementary Material}
\maketitle
\begin{figure*}
    \centering
    \captionsetup{justification=centering}
    \includegraphics[width=0.45\linewidth]{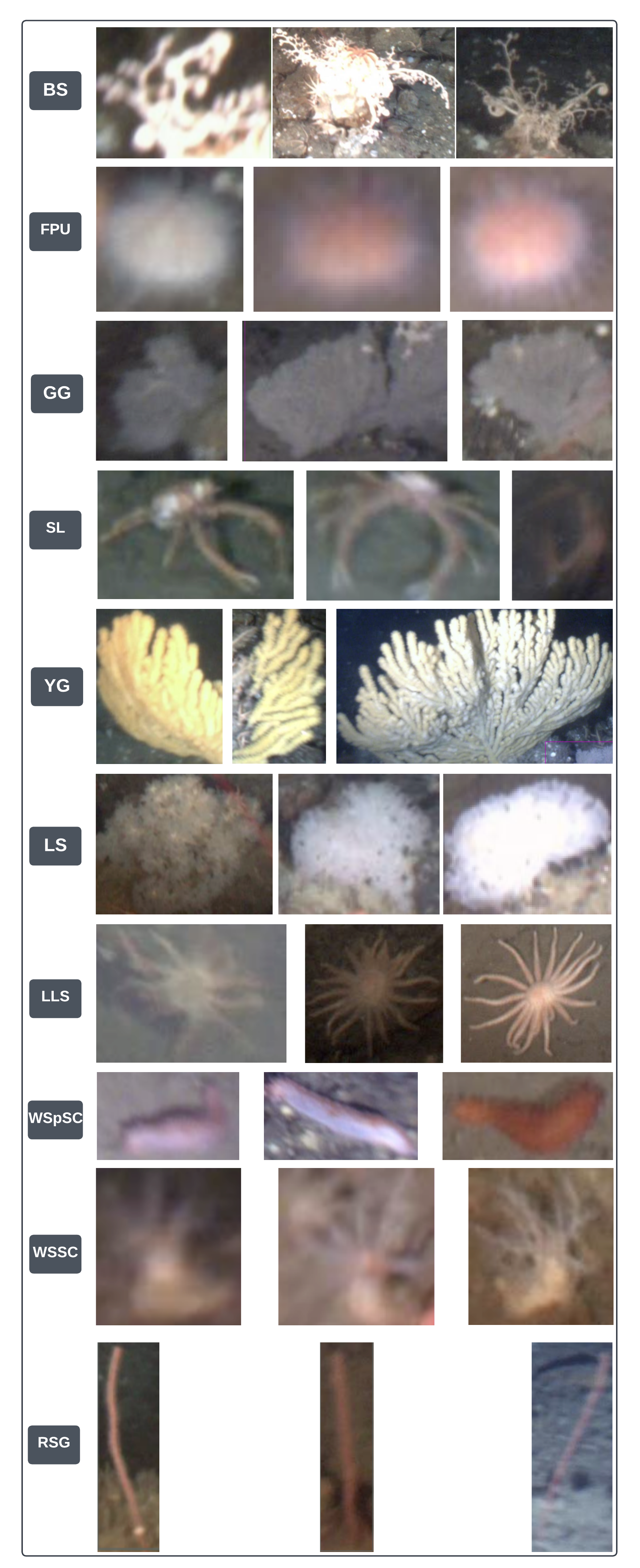}
    \caption{Examples of the ten species that are labelled with bounding boxes in DUSIA. YG stands for yellow gorgonian; BS, basket star; GG, gray gorgonian; LS, laced sponge; WSpSC, white spine sea cucumber; LLS, long-legged sunflower star; SL, squat lobster; FPU, fragile pink urchin; WSSC, white sea slipper cucumber; RSG, red swifita gorgonian.}
    \label{fig:species_appendix}
\end{figure*}

For an example segment of a video from DUSIA, please visit \url{https://youtu.be/dgJnSus2rqI}. Tables \ref{tab:spec_ac_sp} and \ref{tab:count_ac_sp} were taken directly from \cite{mcever2022context}. We include them here for ease of access to this information.
Figure \ref{fig:species_appendix} includes additional examples for each of DUSIA's species.

\begin{table*}[hb!]
\centering
\begin{tabular}{lrrrrrrrrrrr}
\hline
      & \multicolumn{1}{l}{BS} & \multicolumn{1}{l}{FPU} & \multicolumn{1}{l}{GG} & \multicolumn{1}{l}{LLS} & \multicolumn{1}{l}{RSG} & \multicolumn{1}{l}{SL} & \multicolumn{1}{l}{LS} & \multicolumn{1}{l}{WSSC} & \multicolumn{1}{l}{WSpSC} & \multicolumn{1}{l}{YG} & \multicolumn{1}{l}{Total} \\ \hline
Train & 1,247                  & 3,675                   & 3,294                  & 735                      & 775                     & 3,264                  & 1,071                  & 1,397                    & 819                       & 1,024                  & 17,301                    \\
Val   & 61                     & 394                     & 259                    & 20                       & 85                      & 594                    & 91                     & 439                      & 51                        & 38                     & 2,032                     \\
Test  & 124                    & 653                     & 277                    & 61                       & 79                      & 1,181                  & 98                     & 506                      & 28                        & 180                    & 3,187                     \\
Total & 1,432                  & 4,722                   & 3,830                  & 816                      & 939                     & 5,039                  & 1,260                  & 2,342                    & 898                       & 1,242                  & 22,520                    \\ \hline
\end{tabular}
\caption{Distribution of bounding box annotations of each species across splits. Note that one species individual may be annotated with multiple bounding boxes as it occurs across multiple frames.}
\label{tab:spec_ac_sp}
\end{table*}

\begin{table*}[hb!]
\centering
\begin{tabular}{@{}lrrrrrrrrrrr@{}}
\toprule
      & \multicolumn{1}{l}{BS} & \multicolumn{1}{l}{FPU} & \multicolumn{1}{l}{GG} & \multicolumn{1}{l}{LLS} & \multicolumn{1}{l}{RSG} & \multicolumn{1}{l}{SL} & \multicolumn{1}{l}{LS} & \multicolumn{1}{l}{WSSC} & \multicolumn{1}{l}{WSpSC} & \multicolumn{1}{l}{YG} & \multicolumn{1}{l}{Total} \\ \midrule
Train & 292                    & 2,828                   & 398                    & 269                     & 190                     & 1,649                  & 517                      & 832                      & 279                       & 103                    & 7,357                     \\
Val   & 17                     & 154                     & 80                     & 8                       & 19                      & 208                    & 40                       & 164                      & 22                        & 9                      & 721                       \\
Test  & 52                     & 420                     & 78                     & 29                      & 48                      & 742                    & 75                       & 317                      & 17                        & 38                     & 1,816                     \\
Total & 361                    & 3,402                   & 556                    & 306                     & 257                     & 2,599                  & 632                      & 1,313                    & 318                       & 150                    & 9,894                     \\ \bottomrule
\end{tabular}
\caption{Distribution of CABOF labels across DUSIA and its splits. As described in Section \ref{sec:invert_annos}, each species individual is counted only once when it touches the bottom of the frame.}
\label{tab:count_ac_sp}
\end{table*}

\bibliographystyle{IEEEtran}
\bibliography{main}